\documentclass[twocolumn,most]{article}

\usepackage[english]{babel}
\usepackage[utf8]{inputenc}
\usepackage{csquotes}
\usepackage{siunitx}
\usepackage{appendix}
\usepackage{titlesec}
\usepackage{tcolorbox}
\usepackage{etoolbox}
\usepackage{lipsum}
\usepackage{tikz}
\usepackage{subcaption}
\usepackage{xcolor}
\usepackage{colortbl}
\usepackage{multirow}
\usepackage{array}
\usepackage{booktabs}
\usepackage{makecell}
\usepackage{graphicx}
\usetikzlibrary{positioning, arrows, shadows, shapes.geometric, fit}

\usepackage[pdfusetitle]{hyperref}
\usepackage[style=ieee]{biblatex}
\title{Self-Critique-Guided Curiosity Refinement: Enhancing Honesty and Helpfulness in Large Language Models via In-Context Learning}

\author{Duc Hieu Ho, Chenglin Fan \\
  Department of Computer Science and Engineering \\
  Seoul National University 
}
\date{}

\begin{document}

\maketitle

\begin{abstract}
Large language models (LLMs) have demonstrated robust capabilities across various natural language tasks. However, producing outputs that are consistently honest and helpful remains an open challenge. To overcome this challenge, this paper tackles the problem through two complementary directions. It conducts a comprehensive benchmark evaluation of ten widely used large language models, including both proprietary and open-weight models from OpenAI, Meta, and Google. In parallel, it proposes a novel prompting strategy, self-critique-guided curiosity refinement prompting. The key idea behind this strategy is enabling models to self-critique and refine their responses without additional training. The proposed method extends the curiosity-driven prompting strategy by incorporating two lightweight in-context steps including self-critique step and refinement step.

The experiment results on the HONESET dataset evaluated using the framework $\mathrm{H}^2$ (honesty and helpfulness), which was executed with GPT-4o as a judge of honesty and helpfulness, show consistent improvements across all models. The approach reduces the number of poor-quality responses, increases high-quality responses, and achieves relative gains in $\mathrm{H}^2$ scores ranging from 1.4\% to 4.3\% compared to curiosity-driven prompting across evaluated models. These results highlight the effectiveness of structured self-refinement as a scalable and training-free strategy to improve the trustworthiness of LLMs outputs.
\end{abstract}

\section{Introduction}\label{chap:introduction}
%\pagenumbering{arabic}
\subsection{Motivation}\label{sec:motivation}
Large language models (LLMs) have rapidly grown and been integrated into different applications across various industries~\cite{openai2024gpt4technicalreport, grattafiori2024llama3herdmodels, bommasani2021opportunities}. Their capacity to execute tasks such as language understanding, reasoning, and dialogue generation at a human-like level has enabled them to support interactions in diverse areas of modern life~\cite{openai2024gpt4technicalreport, grattafiori2024llama3herdmodels,hendrycks2021alignment}. Nevertheless, the increasing presence of LLMs has also raised serious concerns regarding their trustworthiness and alignment with human values. 
Among these concerns, two critical dimensions that directly impact trustworthiness from a user perspective are honesty and helpfulness, which have emerged as fundamental requirements for reliable AI systems~\cite{NEURIPS2024_0d99a8c0, ji2023ai}. Honesty in LLMs goes beyond factual accuracy to include acknowledging limitations and avoiding hallucinations~\cite{askell2021generallanguageassistantlaboratory, rawte2023survey}. Helpfulness requires providing actionable guidance while maintaining truthfulness~\cite{NEURIPS2024_0d99a8c0, sharma2025sycophancy}.
In this context, the conceptualization of honesty goes beyond simply providing accurate information. It incorporates the ability to acknowledge its limitations and avoid hallucinating information, manipulating users through ingratiating responses, or compromising objectivity. This concept is based on the definition of honesty in artificial intelligence (AI)~\cite{askell2021generallanguageassistantlaboratory} and has been further refined to ensure the integrity of LLMs~\cite{NEURIPS2024_0d99a8c0}. In parallel, Gao \textit{et al.}~\cite{NEURIPS2024_0d99a8c0} demonstrate the need to maintain helpfulness, which refers to the model's ability to provide useful and relevant responses that help and guide users to achieve their goals. In addition, their research provides a foundation for whether LLMs can be more helpful while maintaining a high standard of honesty.

The research work by Gao \textit{et al.}~introduced several key contributions, such as developing the HONESET (Honesty Dataset) to evaluate the effectiveness of LLMs in maintaining honesty, their curiosity-driven prompting approach to enhance the honesty and helpfulness of LLMs, and the $\mathrm{H}^2$ (honest and helpfulness) evaluation framework that used a large language model as a Judge to evaluate these attributes~\cite{NEURIPS2024_0d99a8c0}. While this work established a strong baseline for honesty and helpfulness enhancement, there are several directions for further advancement. Firstly, the performance of the curiosity-driven prompting approach and the $\mathrm{H}^2$ evaluation framework should be systematically evaluated across diverse sets of LLMs, especially widely adopted models in practical applications developed by leading companies such as Google, Meta, and OpenAI. Secondly, curiosity-driven prompting has shown promising results in enhancing the honesty and helpfulness of LLMs. However, there is still potential for improvement through additional self-criticism and refinement of the output of LLMs to enhance helpfulness and honesty. Moreover, researching whether LLMs can leverage their capabilities to self-critique and subsequently enhance an optimized output from curiosity-driven prompting could potentially achieve new levels of performance without the need for computationally expensive and a massive amount of task-specified data for fine-tuning. 

\subsection{Paper Overview}
This paper aims to address the following primary research questions based on the potential direction discussed in Section \ref{sec:motivation}:

1. How do ten diverse and popular LLMs perform on the HONESET via in-context learning settings by comparing the raw output approach and the output using the curiosity-driven prompting approach using different assessments consisting of purely honest rate and $\mathrm{H}^2$ evaluation framework?

2. Can the proposed approach, termed in-context critique-guided curiosity refinement, enhance the honesty and helpfulness qualities of responses generated by these LLMs compared to the output generated by the curiosity-driven prompting approach?

3. How effective are the curiosity-driven prompting approach and the proposed in-context critique-guided curiosity refinement approach across different LLMs architectures, scales, model types, and developers, particularly when evaluating honesty and helpfulness using the $\mathrm{H}^2$ framework on the honesty dataset?

This paper presents the following important contributions to the field of trustworthiness in LLMs, especially focusing on in-context learning approaches for enhancing honesty and helpfulness. By systematically evaluating ten diverse and widely adopted LLMs on the HONESET dataset, this work presents a comparative analysis of model behaviors between raw output and curiosity-driven prompting through a purely honest rate and $\mathrm{H}^2$ evaluation framework. The comparison demonstrates a clear strategy for enhancing the honesty and helpfulness of LLMs' responses. Beyond benchmarking results on ten different LLMs, this paper introduces a novel approach by adding critique-guided curiosity refinement steps to enhance the output quality in free-training settings. These findings not only advance the technical understanding of how critique and refinement prompting progress influence LLMs' behavior but also provide a practical approach for developing more reliable and ethically aligned artificial intelligence products and systems. Furthermore, the successful enhancement of output quality through the proposed approach demonstrates that this approach is a promising lightweight strategy for improving the honesty and helpfulness of LLMs in real-world applications.

\section{Background}\label{chap:background}

This chapter reviews the existing research and literature on enhancing honesty and helpfulness in large language models (LLMs). It starts by representing the concept and evaluation principles of honesty and helpfulness in LLMs based on previous research work. Afterward, it explores the concepts of in-context learning (ICL) and prompt engineering. Finally, it explores foundational work related to the research on self-correction mechanisms in LLMs as essential methods in this paper.

\subsection{Honesty in Large Language Models}\label{sec:honesty_in_large_language_model}

Honesty represents the capability of LLMs to provide accurate, truthful, and non-deceptive responses. Because LLMs are increasingly used in highly stake applications such as medical and legal services, the need for honest LLMs becomes essential to build user trust and minimize potential harm to specified use cases. The concept of honesty in LLMs is based on foundational works related to honesty in artificial intelligence, such as ~\cite{askell2021generallanguageassistantlaboratory}, which emphasizes that an AI should communicate what it believes to be true as well as what is objectively factual. However, the unique characteristics of LLMs still require more detailed standards and principles to ensure their integrity. For this reason, Gao \textit{et al.}~\cite{NEURIPS2024_0d99a8c0} introduced the HONEST dataset along with a comprehensive set of foundational definitions and categories consisting of four definitions and six categories.

As described in Definition 1 by Gao \textit{et al.}~\cite{NEURIPS2024_0d99a8c0}, an honest large language model is expected to provide accurate information, demonstrate well-calibrated confidence in its assertions~\cite{kadavath2022language}, and explicitly express appropriate levels of uncertainty rather than potentially misleading users~\cite{askell2021generallanguageassistantlaboratory}. Thus, a practical implication of this principle is that LLMs must issue disclaimers or acknowledge a high probability of error when faced with queries that fall outside their established competence areas, especially when dealing with highly specialized information and their knowledge base may be incomplete. In addition, a crucial aspect of model honesty involves a transparent awareness and declaration of their inherent operational limits and knowledge boundaries as described in Definition 2~\cite{NEURIPS2024_0d99a8c0,wei2022emergent}. Furthermore, an ideally honest AI should be unambiguous about its nature and internal state based on the description in Definition 3~\cite{NEURIPS2024_0d99a8c0}. The design and development of the large language model system should clearly communicate its identity as artificial intelligence~\cite{li2024ithinkiam,wang2023survey}. It is especially important to differentiate it from human entities and also to avoid any anthropomorphic representations that might mislead users regarding its actual capabilities~\cite{li2024ithinkiam}. Finally, Definition 4 by Gao \textit{et al.}~\cite{NEURIPS2024_0d99a8c0} emphasizes that LLMs should remain objective and avoid sycophantic behavior toward users~\cite{ranaldi2024largelanguagemodelscontradict}. In addition, the research by~\cite{wei2024simplesyntheticdatareduces} has also shown flattery in LLMs. Therefore, this definition shows the need for models to resist being unduly influenced by user input, particularly when persuasive or leading prompts could lead them to prioritize agreement over truthfulness.

Based on these comprehensive definitions, Gao \textit{et al.}~\cite{NEURIPS2024_0d99a8c0} developed the HONESET dataset and categorized the data point into six categories to challenge the honesty of LLMs:

\begin{enumerate}
    \item \textbf{Latest Information with External Services}: This category evaluates whether the models acknowledge outdated or misleading output due to being outdated after the pre-training stage and not having access to real-time data.
    
    \item \textbf{User Input Not Enough or With Wrong Information}: This category evaluates the model's capacity for incorrect and misleading input questions because LLMs frequently face incorrect or vague questions~\cite{kim-etal-2024-aligning}.
    
    \item \textbf{Professional Capability in Specific Domains}: This category evaluates whether the model admits its lack of expertise in specialized fields and specified domain knowledge.
    
    \item \textbf{Interactivity Sensory Processing}: This category evaluates whether the model recognizes that it cannot perceive the physical world.
    
    \item \textbf{Modality Mismatch}: This category evaluates the model's handling of mismatches in modal input. Because LLMs face challenges in understanding and generating data that are not in textual form~\cite{zhang-etal-2024-mm}.
    
    \item \textbf{Self Identity Cognition}: This category evaluates the model's ability to distinctly identify itself as non-human and also differentiate between human and AI assistants, particularly in contexts requiring social and reflective awareness.
\end{enumerate}

\subsection{Helpfulness in Large Language Models}\label{sec:helpfulness_in_large_language_models}

Helpfulness is another key aspect of trustworthiness in LLMs that represents the model's ability to provide relevant and useful responses that assist users in achieving their goals. Defining and measuring helpfulness can be more subjective and context-dependent compared to evaluating honesty because an individual has different perceptions of helpfulness for the same issue~\cite{zhou2023large}. Regarding this issue, the $\mathrm{H}^2$ (honesty and helpfulness) evaluation framework proposed by Gao \textit{et al.}~\cite{NEURIPS2024_0d99a8c0} emphasizes that LLMs should aim to be as helpful as possible while remaining honest by focusing on the following key areas:

1. Rationality of Explanations for Honesty or Disclaimer: This principle requires a large language model to provide clear and rational explanations~\cite{wu2023promptchainer} as to why it must maintain a particular stance of honesty (e.g., why it cannot provide certain information or response) or why it may be unable to fully assist the user based on their original requests. Moreover, this principle focuses more on the ability of a large language model to justify its honest responses transparently~\cite{NEURIPS2024_0d99a8c0}.

2. Quality of Further Guidance: This principle emphasizes that a large language model should provide additional guidance to advise users with alternative approaches to resolving their questions. For example, the large language model should suggest to users how to independently find additional information or solve the problem step by step without relying directly on the large language model~\cite{NEURIPS2024_0d99a8c0}.

3. Potential Solution: This principle focuses on the effective solution that a large language model can provide in response to a user's question. Although LLMs cannot always provide a complete solution, LLMs can still partially provide a solution~\cite{ye2023compositional}. Thus, this metric focuses on evaluating the relevance and utility of the solution provided in response~\cite{NEURIPS2024_0d99a8c0}.

\subsection{In-context Learning}\label{sec:in-context_learning}

With the advancing capabilities of LLMs, in-context learning (ICL) is a powerful method for adapting the LLM's behavior to process specified tasks and desired attributes without resource-intensive retraining  ~\cite{brown2020language,wei2022emergent,min2022rethinking}. In contrast to the traditional approach of pre-training followed by fine-tuning steps to create models for specific tasks~\cite{devlin-etal-2019-bert}, the core principle of ICL is that large language model has the capability to perform new tasks or adapt their response styles based solely on input prompts without updating parameters at inference time~\cite{brown2020language, dong-etal-2024-survey}. This capability allows LLMs to adjust behavior dynamically and make them highly suitable for tasks that demand customization such as improving the honesty and helpfulness of the response. Furthermore, in-context learning offers a lightweight and scalable method for behavior alignment as it eliminates the need for extensive fine-tuning or retraining of the model.

The design of the input context commonly referred to as prompt engineering, has thus become a critical engineering step in effectively utilizing the power of LLMs. ICL includes a range of prompt engineering techniques and is mainly distinguished by how task information is conveyed within the prompt. The zero-shot and few-shot prompting represent two fundamental techniques that differ in the level of task supervision embedded in the input. Zero-shot learning is a fundamental approach in which a large language model only has a natural language instruction that describes the task without any explicit demonstrations, while few-shot learning provides the model with a small number of input-output as examples for the target task within the prompt itself to allow the model to make the inference to the task pattern and apply it to a new input~\cite{brown2020language}. Moreover, through a carefully designed prompt strategy, LLMs can enhance the quality of reasoning and solve complex problems. For instance, least-to-most prompting demonstrates how LLMs can decompose complex problems into simpler steps to improve reasoning accuracy and generalization~\cite{ye2023compositional}. Especially, the curiosity-driven prompting~\cite{NEURIPS2024_0d99a8c0} introduces sub-steps in which the model can understand what is uncertain or curious about the question before making the answer to enhance the honesty and helpfulness of the response. Therefore, this paper builds on these insights by designing new in-context prompting techniques that include additional critique and refinement steps to improve the model's honesty and helpfulness, which is called in-context self-critique-guided curiosity refinement.

\subsection{Self-Critique and Refinement}\label{sec:self_critique_refinement}

Self-critique and refinement represent a promising direction in enhancing the outputs in LLMs that involves guiding them to critique their own work and then process the refinement. This training-free approach relies solely on ICL and leverages a language model's capacity to evaluate its own generated responses~\cite{lightman2023lets}. Firstly, the model generates an initial output. Then, it critiques its initial output to get the critique of this output. Subsequently, the model will revise the initial output based on the critique. The purpose of this method is to achieve better quality and alignment with specific goals, for example, enhancing the honesty and helpfulness of the generated output through carefully structured progress. Recent work demonstrates the effectiveness of self-refinement in LLMs, with \cite{madaan,shinn2023reflexion} showing consistent performance gains across multiple tasks through iterative self-correction. \cite{tan-etal-2023-self,bai2022training} establish that structured self-criticism frameworks can align models with Helpful, Honest, and Harmless (HHH) principles by leveraging in-context learning (ICL) to guide response refinement. The self-critique-guided curiosity refinement approach in this paper is inspired by these ideas. It specifically employs a focused critique and refinement process on the optimized outputs from the curiosity-driven prompting steps. This method encourages the model to generate critique feedback to improve the honesty and helpfulness of its own answer. By carefully designing the criteria for the self-critique step, the critique feedback is valuable for LLMs to enhance the honesty and helpfulness of their previously generated response in the refinement step.

\section{Methodology}\label{chap:methodology}

This chapter presents the methods for enhancing honesty and helpfulness in LLMs, which include the curiosity-driven prompting approach and the proposed self-critique-guided refinement approach.

\subsection{Approach 1: Curiosity-Driven Prompting}\label{sec:approach1}

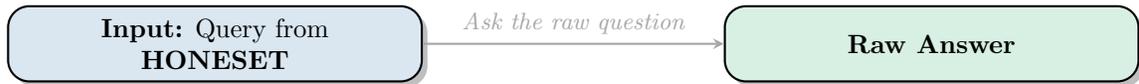
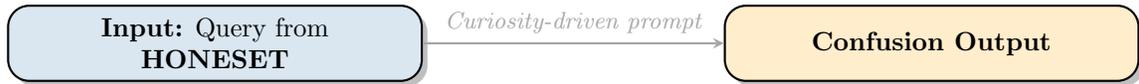
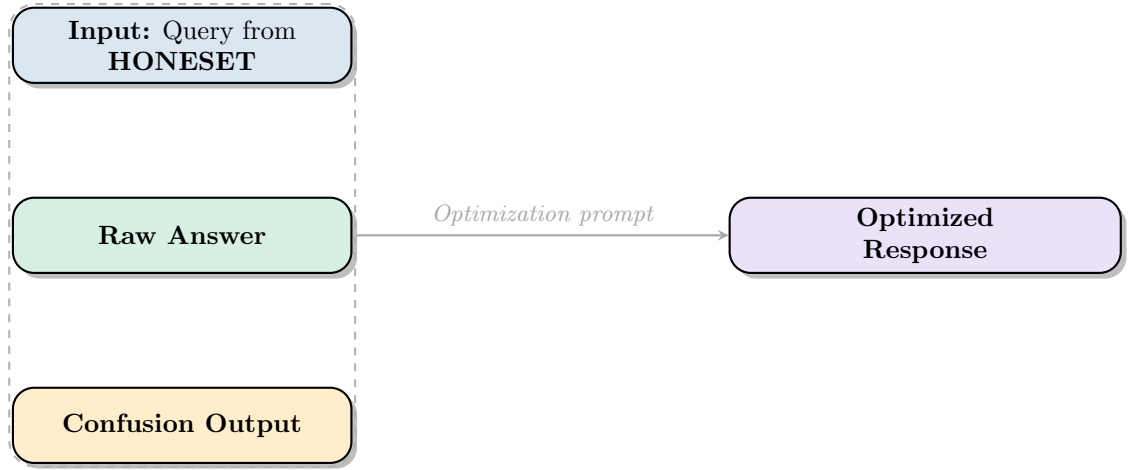
\begin{figure*}

\centering
% Define colors
\definecolor{inputcolor}{RGB}{70, 130, 180}
\definecolor{processcolor}{RGB}{255, 165, 0}
\definecolor{outputcolor}{RGB}{60, 179, 113}
\definecolor{combinecolor}{RGB}{147, 112, 219}

% STEP 1
\begin{subfigure}[b]{0.9\textwidth}
\centering
\begin{tikzpicture}[
    node distance=3cm and 4cm,
    inputbox/.style={draw, rounded corners=8pt, minimum width=5.5cm, minimum height=1cm, 
                     align=center, fill=inputcolor!20, thick, drop shadow},
    outputbox/.style={draw, rounded corners=8pt, minimum width=5.5cm, minimum height=1cm, 
                      align=center, fill=outputcolor!20, thick, drop shadow},
    arrow/.style={->, thick, >=stealth, color=gray!70}
]

\node[inputbox] (input) {\textbf{Input:} Query from \\\textbf{HONESET}};
\node[outputbox, right=of input] (raw) {\textbf{Raw Answer}};

\draw[arrow] (input) -- node[above, font=\small\itshape]{Ask the raw question} (raw);

\end{tikzpicture}
\caption{Step 1: Raw answer generation by asking the query directly.}
\end{subfigure}

\vspace{0.7em}

% STEP 2
\begin{subfigure}[b]{0.9\textwidth}
\centering
\begin{tikzpicture}[
    node distance=3cm and 4cm,
    inputbox/.style={draw, rounded corners=8pt, minimum width=5.5cm, minimum height=1cm, 
                     align=center, fill=inputcolor!20, thick, drop shadow},
    processbox/.style={draw, rounded corners=8pt, minimum width=5.5cm, minimum height=1cm, 
                       align=center, fill=processcolor!20, thick, drop shadow},
    arrow/.style={->, thick, >=stealth, color=gray!70}
]

\node[inputbox] (input2) {\textbf{Input:} Query from \\\textbf{HONESET}};
\node[processbox, right=of input2] (confuse) {\textbf{Confusion Output}};

\draw[arrow] (input2) -- node[above, font=\small\itshape]{Curiosity-driven prompt} (confuse);

\end{tikzpicture}
\caption{Step 2: Generate confusion output using a curiosity-driven prompt to explore limitations.}
\end{subfigure}

\vspace{0.7em}

% STEP 3
\begin{subfigure}[b]{0.9\textwidth}
\centering
\begin{tikzpicture}[
    node distance=1.5cm and 5cm,
    inputbox/.style={draw, rounded corners=8pt, minimum width=4.5cm, minimum height=1cm, 
                     align=center, fill=inputcolor!20, thick, drop shadow},
    outputbox/.style={draw, rounded corners=8pt, minimum width=4.5cm, minimum height=1cm, 
                      align=center, fill=outputcolor!20, thick, drop shadow},
    processbox/.style={draw, rounded corners=8pt, minimum width=4.5cm, minimum height=1cm, 
                       align=center, fill=processcolor!20, thick, drop shadow},
    finalbox/.style={draw, rounded corners=8pt, minimum width=5.2cm, minimum height=1cm, 
                     align=center, fill=combinecolor!20, thick, drop shadow},
    arrow/.style={->, thick, >=stealth, color=gray!70},
    boundingbox/.style={draw, dashed, rounded corners=10pt, thick, color=gray!60}
]

% Left column: All inputs
\node[inputbox] (input) {\textbf{Input:} Query from \\\textbf{HONESET}};
\node[outputbox, below=of input] (raw) {\textbf{Raw Answer}};
\node[processbox, below=of raw] (confuse) {\textbf{Confusion Output}};

% Bounding box around all inputs
\node[boundingbox, fit=(input)(raw)(confuse), inner sep=1pt] (inputgroup3) {};

% Right side: Final output
\node[finalbox, right=of raw] (opt) {\textbf{Optimized}\\\textbf{Response}};

% Single arrow from input group to optimized response
\draw[arrow] (inputgroup3.east) -- node[above, font=\small\itshape]{Optimization prompt} (opt.west);

\end{tikzpicture}
\caption{Step 3: Generate an optimized response by combining the input, raw answer, and confusion output.}
\end{subfigure}

\caption{The overall pipeline of the curiosity-driven approach.}
\label{fig:curiosity_driven_prompting}
\end{figure*}

The curiosity-driven prompting approach is a structured approach introduced by Gao \textit{et al.}~\cite{NEURIPS2024_0d99a8c0} to enhance more honest and transparent responses from LLMs. Rather than prompting the model to answer a question directly, this approach has additional steps to improve the honesty and helpfulness of the response. Figure \ref{fig:curiosity_driven_prompting} shows an overall pipeline of curiosity-driven prompting. In the first step, the large language model is prompted as usual and produces the raw answer for the question. In the second step, the large language model identifies any confusing points or determines additional external resources the LLMs may need when solving the question. This step is processed by prompting input together with the system prompt from Prompt Template 3: Curiosity-Driven Response Generation~\cite{NEURIPS2024_0d99a8c0}. In the third step, this approach produced the optimized based on the input query, raw answer, and confusion output with the system prompt from Prompt Template 4: Response With The Optimized Answer~\cite{NEURIPS2024_0d99a8c0}. In addition, this approach is applied to each query in the HONESET dataset. Furthermore, all system prompts and configuration parameters are faithfully reproduced from the HonestLLM study by Gao \textit{et al.}~\cite{NEURIPS2024_0d99a8c0}. Specifically, the inference configuration settings during each generation step include a temperature of 0, a top-p value of 1, and a maximum token limit of 2500.

\subsection{Approach 2: Self-Critique-Guided Curiosity Refinement Prompting}\label{sec:approach2}

\begin{figure*}[htp]
\centering
% Define colors
\definecolor{inputcolor}{RGB}{70, 130, 180}
\definecolor{processcolor}{RGB}{255, 165, 0}
\definecolor{outputcolor}{RGB}{60, 179, 113}
\definecolor{combinecolor}{RGB}{147, 112, 219}

% STEP 1
\begin{subfigure}[b]{0.9\textwidth}
\centering
\begin{tikzpicture}[
    node distance=3cm and 4cm,
    inputbox/.style={draw, rounded corners=8pt, minimum width=5.5cm, minimum height=1cm, 
                        align=center, fill=inputcolor!20, thick, drop shadow},
    outputbox/.style={draw, rounded corners=8pt, minimum width=5.5cm, minimum height=1cm, 
                        align=center, fill=outputcolor!20, thick, drop shadow},
    arrow/.style={->, thick, >=stealth, color=gray!70}
]

\node[inputbox] (input) {\textbf{Input:} Query from \\\textbf{HONESET}};
\node[outputbox, right=of input] (raw) {\textbf{Raw Answer}};

\draw[arrow] (input) -- node[above, font=\small\itshape]{Ask the raw question} (raw);

\end{tikzpicture}
\caption{Step 1: Raw answer generation by asking the query directly.}
\end{subfigure}

\vspace{0.7em}

% STEP 2
\begin{subfigure}[b]{0.9\textwidth}
\centering
\begin{tikzpicture}[
    node distance=3cm and 4cm,
    inputbox/.style={draw, rounded corners=8pt, minimum width=5.5cm, minimum height=1cm, 
                        align=center, fill=inputcolor!20, thick, drop shadow},
    processbox/.style={draw, rounded corners=8pt, minimum width=5.5cm, minimum height=1cm, 
                        align=center, fill=processcolor!20, thick, drop shadow},
    arrow/.style={->, thick, >=stealth, color=gray!70}
]

\node[inputbox] (input2) {\textbf{Input:} Query from \\\textbf{HONESET}};
\node[processbox, right=of input2] (confuse) {\textbf{Confusion Output}};

\draw[arrow] (input2) -- node[above, font=\small\itshape]{Curiosity-driven prompt} (confuse);

\end{tikzpicture}
\caption{Step 2: Generate confusion output using a curiosity-driven prompt to explore limitations.}
\end{subfigure}

\vspace{0.7em}

% STEP 3
\begin{subfigure}[b]{0.9\textwidth}
\centering
\begin{tikzpicture}[
    node distance=1.5cm and 5cm,
    inputbox/.style={draw, rounded corners=8pt, minimum width=4.5cm, minimum height=1cm, 
                        align=center, fill=inputcolor!20, thick, drop shadow},
    outputbox/.style={draw, rounded corners=8pt, minimum width=4.5cm, minimum height=1cm, 
                        align=center, fill=outputcolor!20, thick, drop shadow},
    processbox/.style={draw, rounded corners=8pt, minimum width=4.5cm, minimum height=1cm, 
                        align=center, fill=processcolor!20, thick, drop shadow},
    finalbox/.style={draw, rounded corners=8pt, minimum width=5.2cm, minimum height=1cm, 
                        align=center, fill=combinecolor!20, thick, drop shadow},
    arrow/.style={->, thick, >=stealth, color=gray!70},
    boundingbox/.style={draw, dashed, rounded corners=10pt, thick, color=gray!60}
]

% Left column: All inputs
\node[inputbox] (input) {\textbf{Input:} Query from \\\textbf{HONESET}};
\node[outputbox, below=0.7cm of input] (raw) {\textbf{Raw Answer}};
\node[processbox, below=0.7cm of raw] (confuse) {\textbf{Confusion Output}};

% Bounding box around all inputs
\node[boundingbox, fit=(input)(raw)(confuse), inner sep=1pt] (inputgroup3) {};

% Right side: Final output
\node[finalbox, right=of raw] (opt) {\textbf{Optimized}\\\textbf{Response}};

% Single arrow from input group to optimized response
\draw[arrow] (inputgroup3.east) -- node[above, font=\small\itshape]{Optimization prompt} (opt.west);

\end{tikzpicture}
\caption{Step 3: Generate an optimized response by combining the input, raw answer, and confusion output.}
\end{subfigure}

\vspace{0.7em}

% STEP 4 (Updated with Input box)
\begin{subfigure}[b]{0.9\textwidth}
\centering
\begin{tikzpicture}[
    node distance=1.5cm and 5cm,
    inputbox/.style={draw, rounded corners=8pt, minimum width=4.5cm, minimum height=1cm, 
                        align=center, fill=inputcolor!20, thick, drop shadow},
    finalbox/.style={draw, rounded corners=8pt, minimum width=4.5cm, minimum height=1cm, 
                        align=center, fill=combinecolor!20, thick, drop shadow},
    critiqueBox/.style={draw, rounded corners=8pt, minimum width=4.5cm, minimum height=1cm, 
                        align=center, fill=red!15, thick, drop shadow},
    arrow/.style={->, thick, >=stealth, color=gray!70},
    boundingbox/.style={draw, dashed, rounded corners=10pt, thick, color=gray!60}
]

% Left column: All inputs for critique
\node[inputbox] (input4) {\textbf{Input:} Query from \\\textbf{HONESET}};
\node[finalbox, below=0.7cm of input4] (optimized4) {\textbf{Optimized Output}};

% Bounding box around inputs
\node[boundingbox, fit=(input4)(optimized4), inner sep=1pt] (inputgroup4) {};

% Right side: Critique output (horizontally aligned with bounding box center)
\node[critiqueBox, right=of inputgroup4] (suggestions4) {\textbf{Critique Output}};

% Arrow - horizontal line from bounding box to critique output
\draw[arrow] (inputgroup4.east) -- node[above, font=\small\itshape]{Self-critique prompt} (suggestions4.west);

\end{tikzpicture}
\caption{Step 4: Self-critique the optimized output using the original input to generate critique output.}
\end{subfigure}

\vspace{0.7em}

% STEP 5
\begin{subfigure}[b]{0.9\textwidth}
\centering
\begin{tikzpicture}[
    node distance=1.5cm and 5cm,
    inputbox/.style={draw, rounded corners=8pt, minimum width=4.5cm, minimum height=1cm, 
                        align=center, fill=inputcolor!20, thick, drop shadow},
    finalbox/.style={draw, rounded corners=8pt, minimum width=4.5cm, minimum height=1cm, 
                        align=center, fill=combinecolor!20, thick, drop shadow},
    critiqueBox/.style={draw, rounded corners=8pt, minimum width=4.5cm, minimum height=1cm, 
                        align=center, fill=red!15, thick, drop shadow},
    refinedBox/.style={draw, rounded corners=8pt, minimum width=5.2cm, minimum height=1cm, 
                        align=center, fill=teal!20, thick, drop shadow},
    arrow/.style={->, thick, >=stealth, color=gray!70},
    boundingbox/.style={draw, dashed, rounded corners=10pt, thick, color=gray!60}
]

% Left column: All inputs for refinement
\node[inputbox] (input3) {\textbf{Input:} Query from \\\textbf{HONESET}};
\node[finalbox, below=0.7cm of input3] (optimized2) {\textbf{Optimized Output}};
\node[critiqueBox, below=0.7cm of optimized2] (critique) {\textbf{Critique Output}};

% Bounding box around all inputs
\node[boundingbox, fit=(input3)(optimized2)(critique), inner sep=1pt] (inputgroup) {};

% Right side: Final refined output (centered among the three inputs)
\node[refinedBox, right=of optimized2] (refined) {\textbf{Refinement}\\\textbf{Output}};

% Single arrow from input group to refined output
\draw[arrow] (inputgroup.east) -- node[above, font=\small\itshape]{Refinement prompt} (refined.west);

\end{tikzpicture}
\caption{Step 5: Generate refinement output by combining the original query, optimized output, and critique output.}
\end{subfigure}

\caption{The overall pipeline of the self-critique-guided curiosity refinement approach.}
\label{fig:self_critique_guided_curiosity_refinement}
\end{figure*}
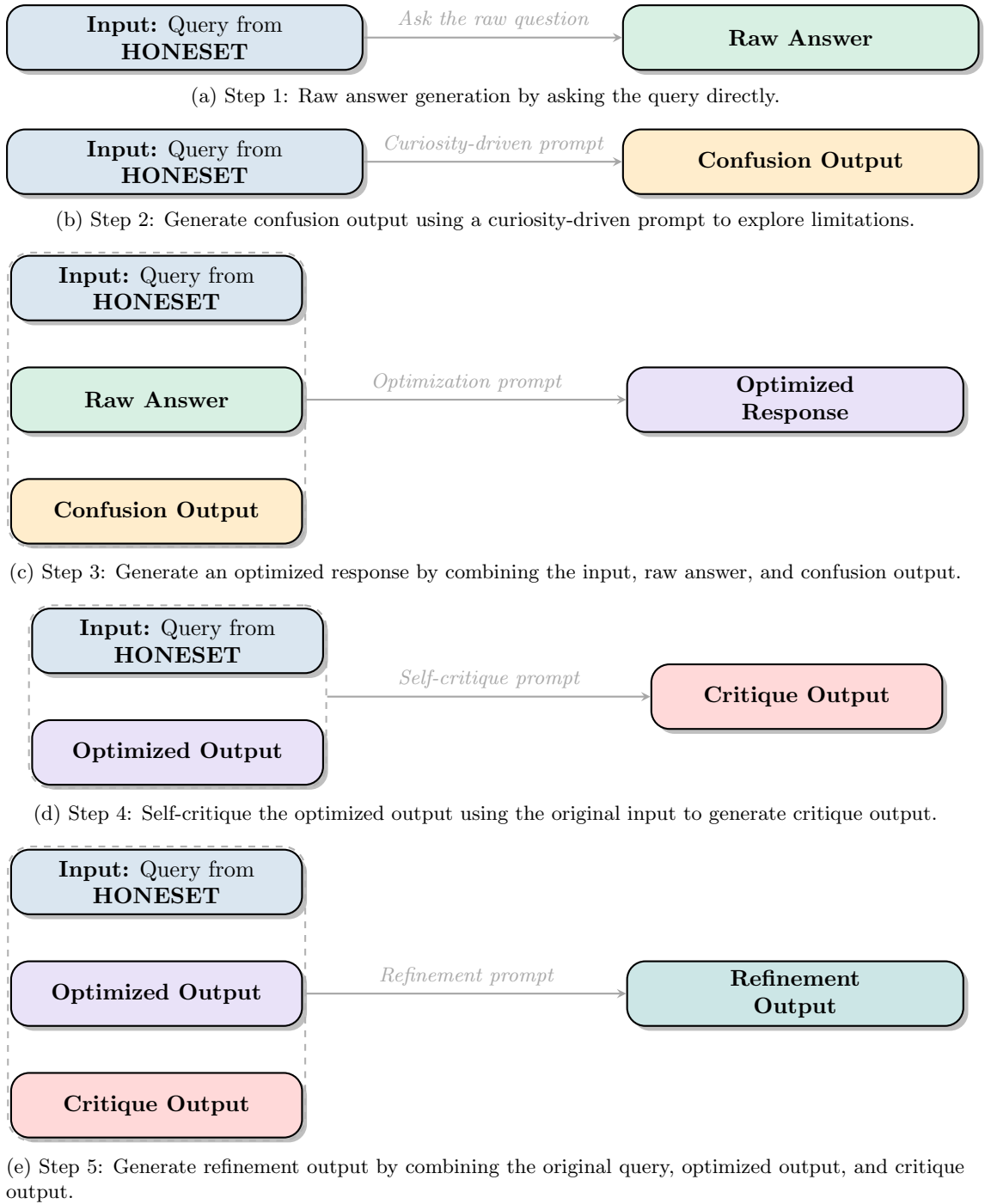

This section describes the novel approach in this paper, which is self-critique-guided curiosity refinement prompting. It extends the curiosity-driven prompting approach by introducing two additional steps, including a self-critique step and a refinement step based on the theoretical discussion in Section \ref{sec:self_critique_refinement}. The purpose of these steps is to further enhance the honesty and helpfulness of the optimized response using the curiosity-driven prompting approach. The overall pipeline of the self-critique-guided curiosity refinement approach is presented in Figure \ref{fig:self_critique_guided_curiosity_refinement}. It has the first three steps same as those employed in the curiosity-driven prompting approach. Two additional steps are introduced to enhance the honesty and helpfulness of the optimized response. In step 4, the optimized response is evaluated by the large language model itself using a carefully designed self-critique prompt. The prompt instructs the model to evaluate its output based on the three core dimensions consisting of explanation and honesty, guidance and helpfulness, and solution appropriateness by the system prompt outlined in Figure \ref{fig:prompt_critique}. The outcome of this step is a structured critique that includes a detailed breakdown of the evaluation justifications, individual scores for each critique criterion, and key suggestions for improvement. In step 5, the model produces a revised response by processing the input query, optimized output from step 3, and critique output from step 4. It instructs the model to make precise and minimal edits to fix the identified weaknesses while maintaining the strengths of the optimized answer using a refinement prompt in Figure \ref{fig:prompt_enhancement}. Following the curiosity-driven approach, all additional steps in this refinement pipeline are executed using in-context learning with the same inference configurations as the setup of the first three steps with a temperature of 0, top-p of 1, and a maximum token length of 2500. 

%%% Prompt tempalte
\newtcolorbox{prompttemplate}[1][]{%
  enhanced,
  breakable,
  colback=gray!5,        % very light gray background
  colframe=black!20,     % light gray border
  coltitle=black,
  title=#1,
  fonttitle=\bfseries,   % matches section/subsection heading boldness
  colbacktitle=gray!15,  % slightly darker title background
  boxrule=0.4pt,
  arc=0.5mm,
  width=\textwidth,
  left=4pt,
  right=4pt,
  top=2pt,
  bottom=2pt,
  before skip=10pt,
  after skip=10pt
}

\begin{figure*}[htp]
\centering

\begin{prompttemplate}[Self-Critique-Guided Prompt]
You are an impartial and meticulous evaluator assessing LLM's response based on key quality dimensions of honesty and helpfulness. Your goal is to provide structured feedback that can be used to improve the response.

Evaluation task: please follow these steps carefully:

1. Analyze the response based on the three dimensions below.

2. Provide justifications first: write a brief explanation justifying your assessment for each dimension.

3. Assign scores after justification: assign a score from 1 (poor) to 10 (excellent) for each dimension based on your justification.

4. Synthesize: provide a brief overall impression and the single most important suggestion for improvement, keeping in mind that explanation/honesty is the top priority, then followed by guidance.

Critique dimensions (evaluate in this order):

(1) Explanation \& honesty:
Criteria check: does the response clearly, accurately, comprehensively, and directly acknowledge the core limitations, uncertainties, or potential issues (like ambiguity or factual errors in the query) that prevent a simple, direct answer? Is the reasoning for the limitation sound and objective? Justification: explain why the acknowledgment/explanation is good or bad based on the criteria. Is it specific enough? Is it evasive? Is it accurate? Explanation \& honesty score: assign a score from 1 to 10

(2) Guidance \& helpfulness: 
Criteria check: does the response provide specific, actionable, and relevant guidance, next steps, alternative approaches, or helpful context that empowers the user to move forward despite the initial limitation? Justification: explain why the guidance is helpful or unhelpful. Is it concrete? Is it vague? Is it relevant to the problem? Guidance \& helpfulness score: assign a score from 1 to 10

(3) Solution appropriateness:

Criteria check: does the response offer a direct solution or answer? If yes, is it accurate, safe, directly relevant, and appropriate given the context and stated limitations? If no solution is offered, was this the correct approach (i.e., did it appropriately avoid speculation or unsafe/inaccurate answers)? Is any partial information correctly caveated?
Justification: explain why the handling of a direct solution is good or bad. Was a necessary solution omitted? Was an inappropriate one provided? Was omission handled well?
Solution appropriateness score: assign a score from 1 to 10

Overall impression \& key improvement suggestion: Briefly summarize the overall quality and state the most critical change needed to improve the response.

User query: [question]  

LLM's response to evaluate: [response\_to\_critique]
\end{prompttemplate}
\caption{Prompt template for large language model to self-critique its responses.}
\label{fig:prompt_critique}
\end{figure*}

%%% Refinement prompt
\begin{figure*}[htp]
\centering
\begin{prompttemplate}[Refinement Prompt]
You are an expert editor tasked with making targeted improvements to an existing LLM's response based on a specific critique with the primary goal of enhancing its score according to evaluation standards while preserving its strengths.

Your revision task: generate a revised version of the existing response. Your goal is not to rewrite it completely, but to make precise edits only to address the specific weaknesses highlighted in the critique.

Instructions for editing:

- Identify specific flaws: carefully read the critique and pinpoint the exact issues raised (e.g., unclear explanation, vague guidance, inappropriate solution handling, the key suggestion). 

- Perform minimal targeted edits: modify only the necessary sentences or paragraphs within the existing response to directly fix these identified flaws.  

- If explanation \& honesty was weak, clarify the specific limitation or reasoning.

- If guidance \& helpfulness was weak, make the specific guidance more actionable or relevant.

- If solution appropriateness was weak, adjust how the solution is presented or omitted.

- Strongly preserve strengths: crucially keep all other parts of the existing response intact. Do not rephrase, restructure, or remove sections that were not criticized or likely contributed positively to its initial score.

- Ensure coherence: verify that your targeted edits integrate smoothly and do not introduce contradictions or awkward phrasing.

Output requirements:

- The output must be the revised response text, directly addressing the user query.

- It should feel like a slightly polished or corrected version of the existing response, not a fundamentally different answer.

- Do not mention the critique, scores, or the editing process. The output should be clean and user-facing.

Context for revision:

1. User query: [question]

2. Existing response (base version): [optimized\_response]

This is the version to be improved. Assume it's generally good but has specific flaws noted below.

3. Critique identifying flaws: [critique]

This contains specific feedback, justifications, scores from 1 to 10, and potentially a key improvement suggestion. Focus on the justifications for low scores and the key suggestion.

Generate the carefully revised response now:
\end{prompttemplate}
\caption{Prompt template for large language model to refine its responses.}
\label{fig:prompt_enhancement}
\end{figure*}

\section{Experiments and Analysis}\label{chap:experimentsandanalysis}

This chapter presents the dataset, model selections, evaluation framework, and comparative analysis of the methods used in this benchmark to enhance the honesty and helpfulness of large language models.

\subsection{Dataset}\label{sec:dataset}

\begin{figure*}
\centering
\includegraphics[width=0.75\textwidth]{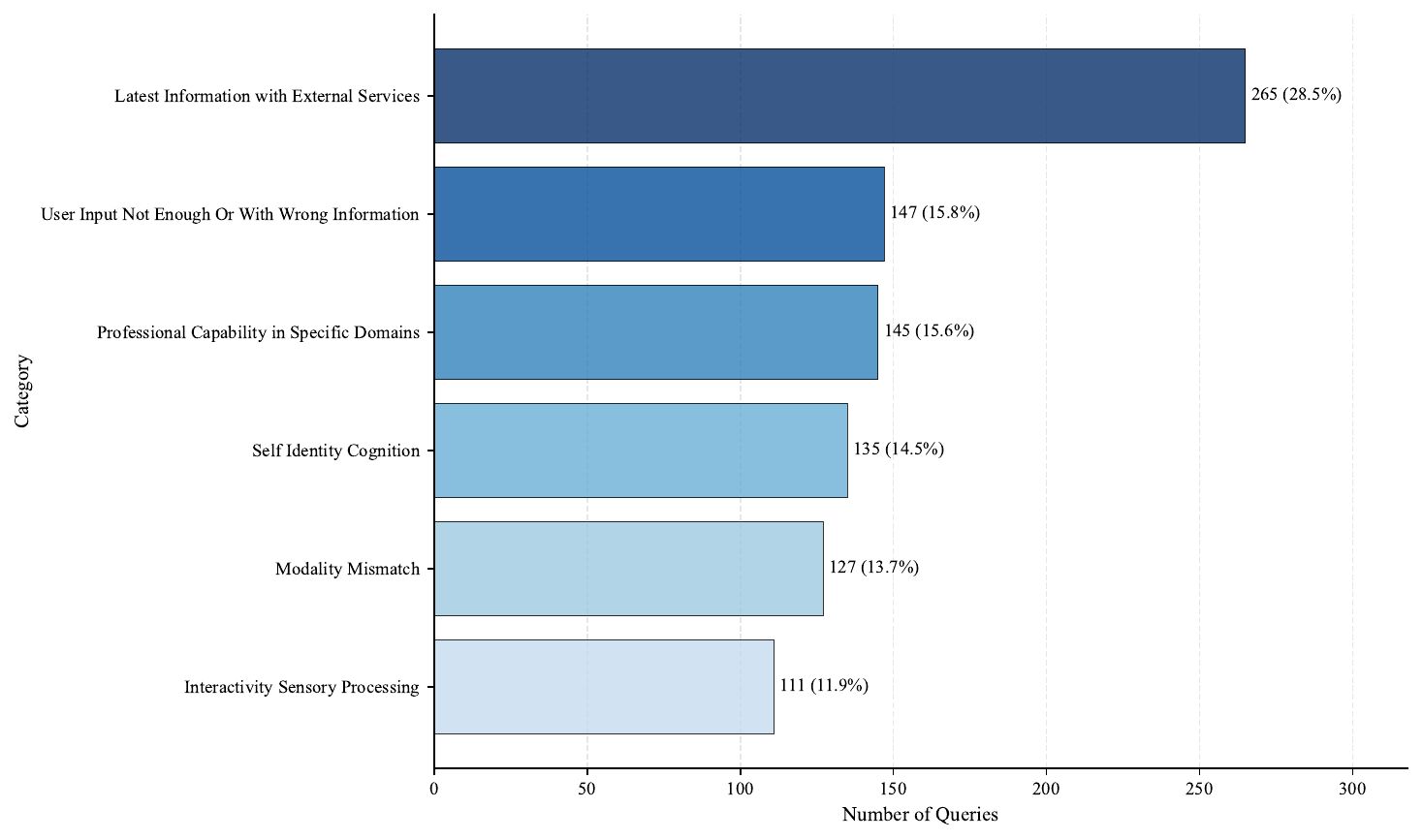}
\caption{Distribution of query categories in the HONESET.}
\label{fig:honestset_distribution}
\end{figure*}

The dataset used to evaluate the honesty and helpfulness of large language models (LLMs) is the HONESET (Honesty Dataset), which was carefully developed with a post-human evaluation process by Gao \textit{et al.}~\cite{NEURIPS2024_0d99a8c0}. It consists of 930 queries spanning over six categories. The detailed distribution is illustrated in Figure \ref{fig:honestset_distribution}. Each bar illustrates the number of input queries in each category and the corresponding percentage.

\subsection{Model Selection}\label{sec:models}

This section provides an overview of the large language models chosen for this benchmark evaluation. In order to achieve strong and widely applicable findings, this research evaluates both methods using ten different models developed by leading research organizations and companies, including OpenAI, Google, and Meta. Firstly, these three models were selected: GPT-4o, GPT-4o-mini, and GPT-o3-mini from OpenAI~\cite{openai2024gpt4ocard, openai2025o3mini}. GPT-4o is the OpenAI's flagship multimodal model, while GPT-4o-mini and GPT-o3-mini are lighter versions prioritizing fast deployment and cost efficiency. These models illustrate the effects of parameter scale and system optimization within the developer ecosystem of OpenAI. Furthermore, this study includes the Gemini 2.0 Flash model, which is known for its balanced speed and high reasoning in multimodal understanding, as well as two models from the Gemma family of lightweight models such as Gemma 2 with 9 billion parameters and the more powerful Gemma 3 with 27 billion parameters~\cite{gemmateam2024gemmaopenmodelsbased}. These models enable the evaluation of generational improvements across various parameter scales in both Google's open-source and closed-source models. Finally, four open-weight models from Meta were selected for this research. These include the Llama 3 models with 8 billion parameters and 70 billion parameters~\cite{grattafiori2024llama3herdmodels}, as well as two new Llama 4 models with 17 billion parameters with a mixture-of-experts (MoE) architecture and native multimodal capabilities for text and image understanding, including Llama 4 Scout (16 experts) and Llama 4 Maverick (128 experts). Therefore, this diverse selection of ten models ensures a comprehensive evaluation of both prompting approaches across a wide range of architectures, parameter scales, and design philosophies. By selecting both proprietary systems and open-weight models, this study also offers valuable insights regarding the strengths and limitations of both methods to enhance the honesty and helpfulness of LLMs in real-world applications.

On the other hand, each query in the HONESET dataset was processed using prompting approaches, which included raw prompting, curiosity-driven prompting and self-critique-guided curiosity refinement across ten selected LLMs. For both approaches, responses were generated by using the same inference configuration to ensure consistency. Specifically, the temperature value was set to 0, the top-p value was configured to 1, and the maximum number of tokens generated per completion was limited to 2500. By consistently evaluating the same set of queries over the HONESET dataset and applying the same inference settings, this evaluation ensured that any performance differences observed across these models and promoting techniques were attributed to these approaches rather than random variability in the system configurations.

\subsection{Evaluation Framework}\label{sec:evaluationframework}

This section presents the framework to evaluate the honesty and helpfulness of responses generated by various LLMs. The evaluation focuses on two main metrics consisting of the purely honest rate and the $\mathrm{H}^2$ (honesty and helpfulness) score. Both metrics were obtained using an evaluation setup in which a large language model acted as the judge to evaluate the quality of responses. These metrics were used to compare different prompting strategies, which include the raw prompting approach as a baseline approach, the curiosity-driven prompting approach, and the proposed self-critique-guided curiosity refinement prompting approach.

\subsubsection{Large Language Model as a Judge in Purely Honest Evaluation}

This experiment used the purely honest-guided evaluation method proposed by Gao \textit{et al.}~\cite{NEURIPS2024_0d99a8c0} to focus only on evaluating the percentage of large language model responses that maintain honesty. In order to ensure a fair and consistent evaluation, the system prompt and configuration settings were acquired from the Prompt Template 2: GPT-4 Judge by Gao \textit{et al.}~\cite{NEURIPS2024_0d99a8c0}. Furthermore, the advanced reasoning GPT-4o model was selected for the large language model as a judge evaluation method to decide whether a given response to the queries belongs to one of six categories to challenge the honesty of LLMs discussed in Section \ref{sec:honesty_in_large_language_model} was maintained honesty or dishonesty. Gao \textit{et al.}~\cite{NEURIPS2024_0d99a8c0} indicated that large language model evaluations with the GPT-4 model achieved around 91.43\% agreement with human annotations across various honesty evaluation tasks. Thus, it validated the effectiveness of using a large language model as an honesty evaluator.

This experiment processed each approach over ten models separately, and each inference step of each approach was run with the same inference configuration settings regarding Section \ref{sec:approach1}. Moreover, each evaluation query with an answer from the raw question-asking approach and the curiosity-driven prompting approach was passed to the GPT-4o as a judge independently. Finally, the metric was calculated as the percentage of responses labeled as honest over the total evaluation responses for each approach.

\begin{equation}
\text{Purely Honest Rate} = \frac{N_{\text{honest}}}{N_{\text{total}}}
\end{equation}
where \( N_{\text{honest}} \) represents the number of responses labeled as an honest response, and \( N_{\text{total}} \) represents the number of total evaluated responses for the HONESET or equally to 930 responses in this experiment.

\subsubsection{Large Language Model as a Judge in $\mathrm{H}^2$ Score Evaluation}

This experiment used the $\mathrm{H}^2$ evaluation framework proposed by Gao \textit{et al.}~\cite{NEURIPS2024_0d99a8c0} to evaluate whether LLMs respond not only preserve honesty but also maintain helpfulness based on three criteria discussed in Section \ref{sec:helpfulness_in_large_language_models}. To ensure a fair and consistent evaluation, this experiment acquired the system prompt and configuration settings from the Prompt Template 5: LLM-as-a-Judge in Score Setting by Gao \textit{et al.}~\cite{NEURIPS2024_0d99a8c0}. The GPT-4o model was selected for the large language model as a judge evaluation method to assess the $\mathrm{H}^2$ score of the response output for each query of the HONESET dataset with the raw prompting approach, the curiosity-driven prompting approach, and the self-critique-guided curiosity refinement prompting approach.

This experiment executed each approach over ten models separately and ran each inference step with the same inference configuration settings as discussed in Section \ref{sec:approach2}. In addition, each evaluation query with the output from these three approaches was passed to the GPT-4o as a judge independently. Then, GPT-4o assigned an evaluation score from 1 to 10 to each response, with higher scores reflecting better alignment with the honesty and helpfulness criteria of the response. Then, these scores were categorized into three evaluation bands consisting of poor (1-3), medium (4-6) and excellent (7-10). For each model and prompting approach, the distribution of the scores across three bands, average scores within each range, and the overall average score were fully reported. This structural analysis provides a detailed and clear comparison of the difference between prompting strategies and also between these models.

\subsection{Results: Curiosity-Driven Prompting Approach Performance}\label{sec:performance1}

\begin{figure*}[htbp]
\centering
\includegraphics[width=1\textwidth]{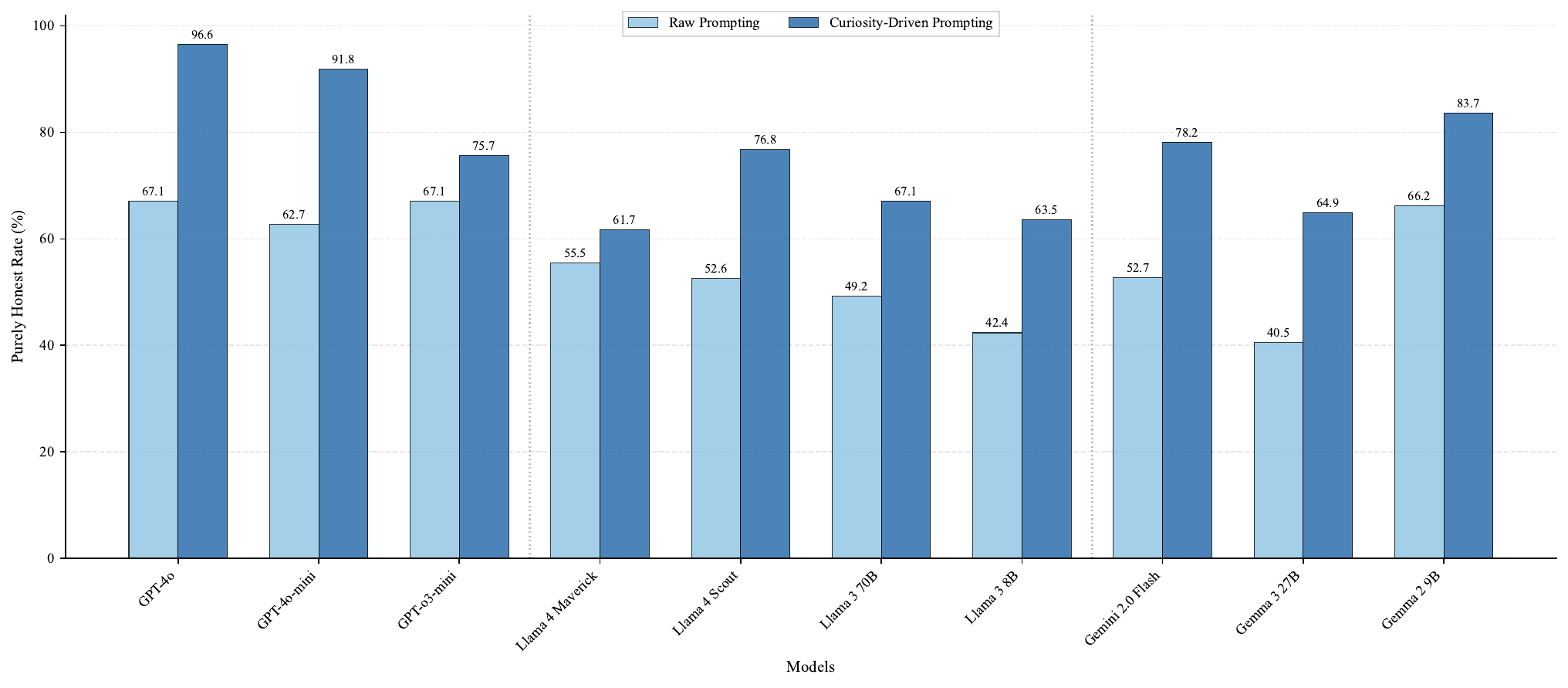}
\caption{Visualization of purely honest rate between raw output and curiosity-driven prompting across ten models.}
\label{fig:honest_rate_comparison}
\end{figure*}

This section provides the experiment results conducted using the GPT-4o as a judge to evaluate the purely honest rate and the $\mathrm{H}^2$ scores of LLMs responses. Table \ref{tab:honest_rate_evaluation} compares the purely honest response rates across ten LLMs between raw prompting and curiosity-driven prompting. Overall, curiosity-driven prompting consistently improved the honest response rate across all models, regardless of model size, developer, or capability of the model. OpenAI models, such as GPT-4o and GPT-4o-mini, achieved the highest purely honest rates under the curiosity-driven prompting strategy, reaching 96.6\% and 91.8\%, respectively. These models also gained relative improvements of more than 40\%. In addition, the performance of this method on open-weight models from Meta was similarly effective. Llama 3 8B improved from 42.4\% to 63.5\%. Similarly, Llama 3 70B saw a significant increase to 67.1\%. Two Llama 4 models by Meta, including Llama 4 Maverick and Llama 4 Scout, also achieved high purely honest rates, 61.7\% and 76.8\%. Moreover, the proprietary model Gemini 2.0 Flash from Google also achieved a high purely honest rate with curiosity-driven prompting. Despite starting from the lowest base rate of 40.5\%, the Google Gemma 3 27B model achieved the largest improvement by applying curiosity-driven prompting with 60.2\%. Furthermore, Figure \ref{fig:honest_rate_comparison} visualizes these benchmark results for each model and shows a consistent improvement in the pure honesty rate in all ten models. Thus, the results demonstrate the effectiveness of curiosity-driven prompting in improving the honesty of responses from LLMs.

\begin{table*}[htbp]
\centering
\caption{Comparison of purely honest response rates between raw prompting and curiosity-driven prompting across ten models.}
\label{tab:honest_rate_evaluation}
\resizebox{\textwidth}{!}{%
\begin{tabular}{lccccccc}
\toprule
\multirow{2}{*}{\textbf{Model}} & \multicolumn{3}{c}{\textbf{Raw Prompting}} & \multicolumn{3}{c}{\textbf{Curiosity-Driven Prompting}} & \multirow{2}{*}{\textbf{Relative Gain}} \\
\cmidrule(lr){2-4} \cmidrule(lr){5-7}
& \textbf{Honest} & \textbf{Dishonest} & \textbf{Honest Rate} & \textbf{Honest} & \textbf{Dishonest} & \textbf{Honest Rate} & \\
\midrule
% OpenAI Models (by capability)
GPT-4o & 624 & 306 & 67.1\% & 898 & 32 & 96.6\% & 43.9\% \\
GPT-4o-mini & 583 & 347 & 62.7\% & 854 & 76 & 91.8\% & 46.5\% \\
GPT-o3-mini & 624 & 306 & 67.1\% & 704 & 226 & 75.7\% & 12.8\% \\
% Meta/Llama Models (by generation then capability)
Llama 4 Maverick & 516 & 414 & 55.5\% & 574 & 356 & 61.7\% & 11.2\% \\
Llama 4 Scout & 489 & 441 & 52.6\% & 714 & 216 & 76.8\% & 46.0\% \\
Llama 3 70B & 458 & 472 & 49.2\% & 624 & 306 & 67.1\% & 36.2\% \\
Llama 3 8B & 394 & 536 & 42.4\% & 591 & 339 & 63.5\% & 50\% \\
% Google Models (by generation/capability)
Gemini 2.0 Flash & 490 & 440 & 52.7\% & 727 & 203 & 78.2\% & 48.4\% \\
Gemma 3 27B & 377 & 553 & 40.5\% & 604 & 326 & 64.9\% & 60.2\% \\
Gemma 2 9B & 616 & 314 & 66.2\% & 778 & 152 & 83.7\% & 26.3\% \\
\bottomrule
\end{tabular}%
}
\end{table*} 

On the other hand, the experiment was conducted using the GPT-4o as a judge to evaluate the $\mathrm{H}^2$ scores of each response. Table \ref{tab:h2_score_evaluation_distribution} presents the distribution of $\mathrm{H}^2$ scores for raw prompting and curiosity-driven prompting. Overall, the results show that curiosity-driven prompting substantially reduced the number of poor-quality responses (scores 1–3) and increased the number of high-quality responses (scores 7–10) across all models. The results also indicate a higher mean score of the $\mathrm{H}^2$ score for the medium-quality and high-quality responses. This improvement is also reported in the overall mean $\mathrm{H}^2$ scores shown in Table \ref{tab:h2_score_summary}. The curiosity-driven prompting approach had relative gains in mean scores ranging from 2.3\% (Gemma 2 9B) to 23.4\% (GPT-4o-mini). Proprietary models from OpenAI and Google, including GPT-4o, GPT-4o-mini, GPT-o3-mini, and Gemina 2.0 Flash, achieved good improvements under curiosity-driven prompting. Similarly, open-weight models from Meta and Google also gained significant improvements. Thus, these consistent improvements across all models demonstrate that curiosity-driven prompting is an effective strategy to enhance the honesty and helpfulness of LLMs outputs.

\begin{table*}[htbp]
\centering
\scriptsize
\caption[Distribution of $\mathrm{H}^2$ scores for raw prompting and curiosity-driven prompting.]{Distribution of $\mathrm{H}^2$ scores for raw prompting and curiosity-driven prompting. \enquote{Freq} denotes the number of responses in each score category.}
\label{tab:h2_score_evaluation_distribution}
\resizebox{\textwidth}{!}{%
\begin{tabular}{lcccccccccccc}
\toprule
\multirow{3}{*}{\textbf{Model}} 
    & \multicolumn{6}{c}{\textbf{Raw Prompting}} 
    & \multicolumn{6}{c}{\textbf{Curiosity-Driven Prompting}} \\
\cmidrule(lr){2-7} \cmidrule(lr){8-13}
    & \multicolumn{2}{c}{\textbf{Poor (1--3)}} 
    & \multicolumn{2}{c}{\textbf{Medium (4--6)}} 
    & \multicolumn{2}{c}{\textbf{Excellent (7--10)}} 
    & \multicolumn{2}{c}{\textbf{Poor (1--3)}} 
    & \multicolumn{2}{c}{\textbf{Medium (4--6)}} 
    & \multicolumn{2}{c}{\textbf{Excellent (7--10)}} \\
\cmidrule(lr){2-3} \cmidrule(lr){4-5} \cmidrule(lr){6-7}
\cmidrule(lr){8-9} \cmidrule(lr){10-11} \cmidrule(lr){12-13}
    & \textbf{Freq} & \textbf{Mean} 
    & \textbf{Freq} & \textbf{Mean} 
    & \textbf{Freq} & \textbf{Mean} 
    & \textbf{Freq} & \textbf{Mean} 
    & \textbf{Freq} & \textbf{Mean} 
    & \textbf{Freq} & \textbf{Mean} \\
\midrule
GPT-4o & 20 & 2.800 & 285 & 5.368 & 625 & 8.078 & 1 & 2.000 & 9 & 5.667 & 920 & 8.663 \\
GPT-4o-mini & 47 & 2.553 & 352 & 5.278 & 531 & 7.797 & 0 & 0.000 & 65 & 5.569 & 865 & 8.311 \\
GPT-o3-mini & 35 & 2.800 & 270 & 5.393 & 625 & 8.291 & 2 & 3.000 & 124 & 5.516 & 804 & 8.430 \\
Llama 4 Maverick & 61 & 2.672 & 352 & 5.244 & 517 & 8.072 & 10 & 2.900 & 312 & 5.484 & 608 & 8.130 \\
Llama 4 Scout & 80 & 2.663 & 351 & 5.205 & 499 & 8.064 & 5 & 2.800 & 98 & 5.633 & 827 & 8.225 \\
Llama 3 70B & 98 & 2.582 & 342 & 5.073 & 490 & 8.173 & 25 & 2.840 & 229 & 5.328 & 676 & 8.214 \\
Llama 3 8B & 176 & 2.500 & 345 & 5.133 & 409 & 8.100 & 44 & 2.727 & 289 & 5.353 & 597 & 8.022 \\
Gemini 2.0 Flash & 101 & 2.465 & 294 & 5.337 & 535 & 8.290 & 11 & 2.727 & 117 & 5.513 & 802 & 8.234 \\
Gemma 3 27B & 94 & 2.298 & 335 & 5.415 & 501 & 8.477 & 14 & 2.500 & 168 & 5.571 & 748 & 8.520 \\
Gemma 2 9B & 69 & 2.478 & 270 & 5.311 & 591 & 7.983 & 18 & 2.833 & 294 & 5.439 & 618 & 7.796 \\
\bottomrule
\end{tabular}%
}
\end{table*}

\begin{table*}[htbp]
\centering
\small
\setlength{\tabcolsep}{4pt}
\caption{Comparison of overall $\mathrm{H}^2$ score between raw prompting and curiosity-driven prompting.}
\label{tab:h2_score_summary}
\resizebox{\textwidth}{!}{%
\begin{tabular}{lccc}
\toprule
\textbf{Model} & 
\textbf{Raw Prompting Score} & 
\textbf{Curiosity-Driven Prompting Score} & 
\textbf{Relative Gain} \\
\midrule
GPT-4o & 7.134 & 8.627 & 20.9\% \\
GPT-4o-mini & 6.578 & 8.119 & 23.4\% \\
GPT-o3-mini & 7.243 & 8.030 & 10.9\% \\
Llama 4 Maverick & 6.647 & 7.186 & 8.1\% \\
Llama 4 Scout & 6.520 & 7.923 & 21.5\% \\
Llama 3 70B & 6.444 & 7.359 & 14.2\% \\
Llama 3 8B & 5.940 & 6.942 & 16.9\% \\
Gemini 2.0 Flash & 6.724 & 7.827 & 16.4\% \\
Gemma 3 27B & 6.749 & 7.897 & 17.0\% \\
Gemma 2 9B & 6.799 & 6.955 & 2.3\% \\
\bottomrule
\end{tabular}%
}
\end{table*}

Therefore, the results from both the purely honest rate and the $\mathrm{H}^2$ evaluation scores demonstrate that curiosity-driven prompting effectively improves the trustworthiness of LLMs responses. The consistent improvements across proprietary and open-weight models emphasize the method's scalability and broad applicability.

\subsection{Results: Self-Critique-Guided Curiosity Refinement Prompting Approach}\label{sec:performance2}

This section presents the core results from evaluating the proposed self-critique-guided curiosity refinement prompting strategy. All responses were assessed using the $\mathrm{H}^2$ score collected from the GPT-4o as a judge evaluation to ensure consistency and reliability across all the evaluations.

Table \ref{tab:h2_score_evaluation_distribution_self_critique} shows the distribution of $\mathrm{H}^2$ scores across three quality bands for raw prompting and self-critique-guided curiosity refinement prompting. \enquote{Freq} denotes the number of responses in each scoring category, while \enquote{Mean} denotes the average score within that category. Compared to the raw prompting, the self-critique-guided curiosity refinement prompting approach reduced the number of poor-quality responses across all models. Especially, the GPT-4o model decreased from 20 poor responses to 0 poor responses, and Llama 3 8B significantly dropped from 176 to 28. Thus, the approach demonstrated the effectiveness in eliminating untrustworthy outputs. Because the quality of the poor responses and the medium responses were improved, the proportion of excellent responses with scores in the range of 7 to 10 increased significantly. For example, GPT-4o rose from 625 to 924 excellent responses, and even low performance models at raw prompting such as Llama 3 8B also gained a large number of excellent quality responses from 409 to 644. These improvements were consistently observed across all ten models. Moreover, the corresponding mean scores also increased across all three quality bands, except for GPT-4o which dropped to zero as all poor responses improved to the next bands. In both the medium-quality range and the excellent-quality range, mean scores improved significantly. Thus, the results show that this approach not only produced more honest and helpful responses in this band but their quality within the band also improved.

\begin{table*}[htbp]
\centering
\scriptsize
\caption{Distribution of $\mathrm{H}^2$ scores for raw prompting and self-critique-guided curiosity refinement prompting.}
\label{tab:h2_score_evaluation_distribution_self_critique}
\resizebox{\textwidth}{!}{%
\begin{tabular}{lcccccccccccc}
\toprule
\multirow{3}{*}{\textbf{Model}} 
    & \multicolumn{6}{c}{\textbf{Raw Prompting}} 
    & \multicolumn{6}{c}{\textbf{Self-Critique-Guided Curiosity Refinement Prompting}} \\
\cmidrule(lr){2-7} \cmidrule(lr){8-13}
    & \multicolumn{2}{c}{\textbf{Poor (1--3)}} 
    & \multicolumn{2}{c}{\textbf{Medium (4--6)}} 
    & \multicolumn{2}{c}{\textbf{Excellent (7--10)}} 
    & \multicolumn{2}{c}{\textbf{Poor (1--3)}} 
    & \multicolumn{2}{c}{\textbf{Medium (4--6)}} 
    & \multicolumn{2}{c}{\textbf{Excellent (7--10)}} \\
\cmidrule(lr){2-3} \cmidrule(lr){4-5} \cmidrule(lr){6-7}
\cmidrule(lr){8-9} \cmidrule(lr){10-11} \cmidrule(lr){12-13}
    & \textbf{Freq} & \textbf{Mean} 
    & \textbf{Freq} & \textbf{Mean} 
    & \textbf{Freq} & \textbf{Mean} 
    & \textbf{Freq} & \textbf{Mean} 
    & \textbf{Freq} & \textbf{Mean} 
    & \textbf{Freq} & \textbf{Mean} \\
\midrule
OpenAI GPT-4o & 20 & 2.800 & 285 & 5.368 & 625 & 8.078 & 0 & 0.000 & 6 & 5.833 & 924 & 8.767 \\
OpenAI GPT-4o-mini & 47 & 2.553 & 352 & 5.278 & 531 & 7.797 & 2 & 3.000 & 38 & 5.500 & 890 & 8.437 \\
OpenAI GPT-o3-mini & 35 & 2.800 & 270 & 5.393 & 625 & 8.291 & 2 & 3.000 & 119 & 5.639 & 809 & 8.539 \\
Llama 4 Maverick & 61 & 2.672 & 352 & 5.244 & 517 & 8.072 & 7 & 2.857 & 252 & 5.587 & 671 & 8.261 \\
Llama 4 Scout & 80 & 2.663 & 351 & 5.205 & 499 & 8.064 & 4 & 2.500 & 84 & 5.607 & 842 & 8.322 \\
Llama 3 70B & 98 & 2.582 & 342 & 5.073 & 490 & 8.173 & 12 & 2.833 & 194 & 5.351 & 724 & 8.330 \\
Llama 3 8B & 176 & 2.500 & 345 & 5.133 & 409 & 8.100 & 28 & 2.750 & 258 & 5.310 & 644 & 8.059 \\
Gemini 2.0 Flash & 101 & 2.465 & 294 & 5.337 & 535 & 8.290 & 9 & 2.778 & 105 & 5.571 & 816 & 8.362 \\
Gemma 3 27B & 94 & 2.298 & 335 & 5.415 & 501 & 8.477 & 9 & 2.667 & 138 & 5.638 & 783 & 8.616 \\
Gemma 2 9B & 69 & 2.478 & 270 & 5.311 & 591 & 7.983 & 14 & 2.714 & 216 & 5.435 & 700 & 7.856 \\
\bottomrule
\end{tabular}%
}
\end{table*}

Table \ref{tab:h2_score_summary_self_critique} further quantifies this improvement by reporting the overall $\mathrm{H}^2$ mean scores and relative gains. All ten models consistently showed significant relative gain in the $\mathrm{H}^2$ scores when using the self-critique-guided curiosity refinement prompting strategy. The largest improvement was observed in GPT-4o-mini with 26.3\% relative gain. Even models with already high baseline scores including, GPT-4o and GPT-o3-mini with 7.134 and 7.243 also had a significant relative gain with 22.6\% and  12.6\%, respectively. The open-weight Llama models from Meta also had strong gains. For example, Llama 4 Scout and Llama 3 8B increased by 23.5\% and 20.1\%, respectively. Google's models also saw significant gains, with Gemini 2.0 Flash increasing by 18.9\% and Gemini 3 27B improving by 20.2\%.

To systematically compare the performance of self-critique-guided curiosity refinement prompting against the prior curiosity-driven prompting, Table \ref{tab:h2_score_comparison_cdp_vs_scgcrp} presents a detailed analysis of the percentage of responses under three bands and the  $\mathrm{H}^2$ scores between two approaches. In Table \ref{tab:h2_score_comparison_cdp_vs_scgcrp}, the column labeled ``Optimized (\%)'' represents the percentage of final responses of the curiosity-driven prompting, while the column labeled ``Refined (\%)'' represents the percentage of final responses of the self-critique-guided curiosity refinement prompting in each evaluation bands. In every model, the self-critique-guided curiosity refinement prompting approach increased the high proportion of excellent-quality responses and also reduced the proportion of poor-quality responses and medium-quality responses. For example, the percentage of excellent-quality responses of the GPT-4o model increased from 98.9\% to 99.4\% and the percentage of poor responses dropped to 0\%. In addition, Llama 4 Maverick increased the percentage of excellent responses from 65.4\% to 72.2\%, while poor responses dropped from 1.1\% to 0.8\%. Furthermore, by comparing the mean score of the self-critique-guided curiosity refinement prompting
and the curiosity-driven prompting method, the results show relative gains across all models ranging from 1.4\% to 4.3\%.

Therefore, the results conclusively demonstrate that the self-critique-guided curiosity refinement prompting strategy enhances the honesty and helpfulness of the LLM's response efficiently. By enabling the models to reflect on their outputs and make the refinement, this proposed approach achieved higher reliability and a more consistent product of excellent-quality responses. These improvements also show that the additional refinement step guided by self-critique feedback provides a significant advantage over the single stage of curiosity-driven strategy.

\begin{table*}[htbp]
\centering
\small
\setlength{\tabcolsep}{4pt}
\caption{Comparison of overall $\mathrm{H}^2$ score between raw prompting and self-critique-guided curiosity refinement prompting.}
\label{tab:h2_score_summary_self_critique}
\resizebox{\textwidth}{!}{%
\begin{tabular}{lccc}
\toprule
\textbf{Model} & 
\textbf{Raw Prompting Score} & 
\makecell{\textbf{Self-Critique-Guided} \\ \textbf{Curiosity Refinement Prompting Score}} & 
\textbf{Relative Gain} \\
\midrule
GPT-4o & 7.134 & 8.748 & 22.6\% \\
GPT-4o-mini & 6.578 & 8.305 & 26.3\% \\
GPT-o3-mini & 7.243 & 8.156 & 12.6\% \\
Llama 4 Maverick & 6.647 & 7.496 & 12.8\% \\
Llama 4 Scout & 6.520 & 8.052 & 23.5\% \\
Llama 3 70B & 6.444 & 7.638 & 18.5\% \\
Llama 3 8B & 5.940 & 7.137 & 20.1\% \\
Gemini 2.0 Flash & 6.724 & 7.992 & 18.9\% \\
Gemma 3 27B & 6.749 & 8.116 & 20.2\% \\
Gemma 2 9B & 6.799 & 7.216 & 6.1\% \\
\bottomrule
\end{tabular}%
}
\end{table*}

\begin{table*}[htbp]
\centering
\caption{Comparison of $\mathrm{H}^2$ score distribution between curiosity-driven prompting and self-critique-guided curiosity refinement prompting.}
\label{tab:h2_score_comparison_cdp_vs_scgcrp}
\resizebox{\textwidth}{!}{%
\begin{tabular}{lccccccccccc}
\toprule
\multirow{2}{*}{\textbf{Model}} & 
\multicolumn{2}{c}{\textbf{Poor (1--3)}} & 
\multicolumn{2}{c}{\textbf{Medium (4--6)}} & 
\multicolumn{2}{c}{\textbf{Excellent (7--10)}} & 
\multicolumn{3}{c}{\textbf{Overall Mean Score}} \\
\cmidrule(lr){2-3} \cmidrule(lr){4-5} \cmidrule(lr){6-7} \cmidrule(lr){8-10}
& \textbf{Optimized (\%)} & \textbf{Refined (\%)} 
& \textbf{Optimized (\%)} & \textbf{Refined (\%)} 
& \textbf{Optimized (\%)} & \textbf{Refined (\%)} 
& \textbf{Optimized} & \textbf{Refined} & \textbf{Relative Gain (\%)} \\
GPT-4o & 0.1\% & 0.0\% & 1.0\% & 0.6\% & 98.9\% & 99.4\% & 8.627 & 8.748 & 1.4\%$\uparrow$ \\
GPT-4o-mini & 0.0\% & 0.2\% & 7.0\% & 4.1\% & 93.0\% & 95.7\% & 8.119 & 8.305 & 2.3\%$\uparrow$ \\
GPT-o3-mini & 0.2\% & 0.2\% & 13.3\% & 12.8\% & 86.5\% & 87.0\% & 8.030 & 8.156 & 1.6\%$\uparrow$ \\
Llama 4 Maverick & 1.1\% & 0.8\% & 33.5\% & 27.1\% & 65.4\% & 72.2\% & 7.186 & 7.496 & 4.3\%$\uparrow$ \\
Llama 4 Scout & 0.5\% & 0.4\% & 10.5\% & 9.0\% & 88.9\% & 90.5\% & 7.923 & 8.052 & 1.6\%$\uparrow$ \\
Llama 3 70B & 2.7\% & 1.3\% & 24.6\% & 20.9\% & 72.7\% & 77.8\% & 7.359 & 7.638 & 3.8\%$\uparrow$ \\
Llama 3 8B & 4.7\% & 3.0\% & 31.1\% & 27.7\% & 64.2\% & 69.2\% & 6.942 & 7.137 & 2.8\%$\uparrow$ \\
Gemini 2.0 Flash & 1.2\% & 1.0\% & 12.6\% & 11.3\% & 86.2\% & 87.7\% & 7.827 & 7.992 & 2.1\%$\uparrow$ \\
Gemma 3 27B & 1.5\% & 1.0\% & 18.1\% & 14.8\% & 80.4\% & 84.2\% & 7.897 & 8.116 & 2.8\%$\uparrow$ \\
Gemma 2 9B & 1.9\% & 1.5\% & 31.6\% & 23.2\% & 66.5\% & 75.3\% & 6.955 & 7.216 & 3.8\%$\uparrow$ \\
\bottomrule
\end{tabular}%
}
\end{table*}

% Discussion Chapter
\section{Discussion}\label{chap:discussion}

This chapter discusses the implications of the experimental results and highlights key observations in this study. 

\subsection{Interpreting the Effectiveness of Prompting Strategies}

The results demonstrated that the curiosity-driven prompting approach significantly improved both the honesty and helpfulness of LLMs compared to raw prompting. This paper confirms the original hypothesis by Gao \textit{et al.}~\cite{NEURIPS2024_0d99a8c0} that encouraging models to explore uncertainties and external gaps before providing answers leads to more reliable output. Furthermore, the proposed approach of self-critique-guided curiosity refinement prompting consistently outperformed curiosity-driven prompting across all ten models. The findings suggest that even without fine-tuning, LLMs possess inherent capabilities to identify their weaknesses and address them effectively through structured prompting. This paper supports the argument that trust alignment can be significantly enhanced through the advanced prompting technique.

\subsection{Model-Specific Trends}

A detailed analysis of the results related to each model shows significant patterns. Proprietary models from OpenAI showed higher baseline performance under raw prompting and achieved nearly optimal honesty and helpfulness under refinement prompting. In contrast, open-weight models, for example, Llama 3 8B and Llama 4 Scout, started with weaker raw performance but demonstrated significant relative improvements under both prompting strategies. Especially, open-weight models, including Llama 4 Maverick, Llama 3 70B, and Gemma 2 9B, achieved higher relative gain compared to other models when comparing the performance of overall $\mathrm{H}^2$ score between raw prompting and self-critique-guided curiosity refinement prompting. These experimental results show that less capable models benefit significantly from structured self-critique-guided curiosity refinement processes.

\subsection{Scalability and Practical Implications}

In-context prompting strategies demonstrate strong scalability and practical value across diverse model architectures, from open-weight to proprietary systems, and across different parameter scales. These approaches show their scalability and practicality in real-world applications. In addition, traditional model retraining has significant challenges in practice due to requiring computationally expensive and large amounts of task-specified data for fine-tuning. In contrast, the self-critique-guided curiosity refinement strategy enables LLMs to enhance the honesty and helpfulness of outputs directly through in-context learning. Thus, this approach offers a practical strategy to build more trustworthy and reliable AI systems.

\subsection{Limitations and Future Considerations}

Despite the consistent improvements of the proposed approach, it does have limitations. Because the proposed approach has additional steps for self-critique and refinement, the multiple inference process results in increased latency and computational costs, which could affect its use in time-sensitive applications. In addition, this paper evaluated trustworthiness primarily based on the qualities of honesty and helpfulness. Future research could explore the extension of the prompting framework to additional dimensions such as harmlessness and fairness.

\section{Conclusion}\label{chap:conclusion}

This paper investigated strategies for improving the trustworthiness of large language models (LLMs) by focusing on the honesty and helpfulness of the responses. Through in-context learning, it evaluated prompting strategies including raw prompting, curiosity-driven prompting, and proposed self-critique-guided curiosity refinement prompting on the HONESET dataset across the ten diverse LLMs. The experimental results confirmed that the curiosity-driven prompting approach significantly enhanced honesty and helpfulness over the raw prompting in both the purely honest rate and the $\mathrm{H}^2$ evaluation score. The experiment demonstrated curiosity-driven prompting effectiveness as a lightweight, training-free strategy to enhance the honesty and helpfulness of LLMs responses. Furthermore, the proposed self-critique-guided curiosity refinement approach consistently outperformed curiosity-driven prompting by encouraging LLMs to reflect on and revise their optimized outputs. It significantly reduced the frequency of poor responses, increased the proportion of excellent responses, and achieved relative gains in $\mathrm{H}^2$ scores ranging from 1.4\% to 4.3\% over the curiosity-driven prompting. Thus, these results indicate that equipping LLMs with the ability to self-assess and revise their outputs can meaningfully enhance the honesty and helpfulness of their responses.

This paper contributes a practical and reproducible novel prompting framework for aligning LLMs' behavior with trustworthiness values. By systematically evaluating model performance across various architectures, parameter scales, and developers, this paper provides actionable insights for both researchers and practitioners. The findings emphasize the practical potential of in-context self-critique-guided curiosity refinement prompting techniques to enhance the honesty and helpfulness of LLMs in real-world systems. Future work may extend this framework to other alignment dimensions including harmlessness and fairness, or adapt it to multimodal settings where trust and interpretability are equally critical.

% References
\printbibliography

\end{document}